%% file: top.tex
\documentclass[10pt,twocolumn,letterpaper]{article}

\usepackage{cvpr}
\usepackage{times}
\usepackage{epsfig}
\usepackage{graphicx}
\usepackage{amsmath}
\usepackage{amssymb}
\usepackage{multirow}
\usepackage[ruled,noend,vlined]{algorithm2e} %
\usepackage{arydshln} %

\usepackage[utf8]{inputenc}
\usepackage[T1]{fontenc}

\usepackage[pagebackref=true,breaklinks=true,letterpaper=true,colorlinks,bookmarks=false]{hyperref}

\cvprfinalcopy %

\def\cvprPaperID{2355} %

\ifcvprfinal\fi
\begin{document}

\title{Learning Monocular 3D Human Pose Estimation from Multi-view Images} %

\author{Helge  Rhodin$^1$ \and Jörg  Spörri$^{2,4}$ \and Isinsu  Katircioglu$^1$ \and Victor  Constantin$^1$ \and Frédéric  Meyer$^3$ \and Erich  Müller$^4$ \and Mathieu  Salzmann$^1$ \and Pascal  Fua$^1$ \vspace{0.1cm} \and
	$^1$CVLab, EPFL, Lausanne, Switzerland \and $^2$Balgrist University Hospital, Zurich, Switzerland \and $^3$UNIL, Lausanne, Switzerland \and $^4$University of Salzburg, Salzburg, Austria
}

\maketitle

\input{tex/defs.tex}

\input{tex/abstract.tex}

\input{tex/intro.tex}

\input{tex/related.tex}
\input{tex/method.tex}

\input{tex/evaluation.tex}

\input{tex/discussion_conclusion.tex}

\section*{Acknowledgement}
This work was supported in part by a Microsoft Joint Research Project.

{\small
\bibliographystyle{ieee}
\bibliography{../../../../bibtex/string,../../../../bibtex/vision,../../../../bibtex/learning,../../../../bibtex/misc,../../../../bibtex/biomed}
}

\end{document}

%% file: tex/defs.tex
\definecolor{olive}{RGB}{50,150,50}

\newif\ifdraft

\ifdraft
 \newcommand{\PF}[1]{{\color{red}{\bf pf: #1}}}
 \newcommand{\pf}[1]{{\color{red} #1}}
 \newcommand{\HR}[1]{{\color{blue}{\bf hr: #1}}}
 \newcommand{\hr}[1]{{\color{blue} #1}}
 \newcommand{\VC}[1]{{\color{green}{\bf vc: #1}}}
  \newcommand{\vc}[1]{{\color{green} #1}}
 \newcommand{\ms}[1]{{\color{olive}{#1}}}
 \newcommand{\MS}[1]{{\color{olive}{\bf ms: #1}}}
 \newcommand{\JS}[1]{{\color{cyan}{\bf js: #1}}}
 \newcommand{\NEW}[1]{{\color{red}{#1}}}

\else
 \newcommand{\PF}[1]{{\color{red}{}	}}
 \newcommand{\pf}[1]{ #1 }
 \newcommand{\HR}[1]{{\color{blue}{}}}
 \newcommand{\hr}[1]{ #1 }
 \newcommand{\VC}[1]{{\color{green}{}}}
 \newcommand{\ms}[1]{ #1 }
 \newcommand{\MS}[1]{{\color{olive}{}}}
 \newcommand{\NEW}[1]{ #1 }
\fi

\newcommand{\TODO}[1]{\textcolor{red}{#1}}
\newcommand{\R}{\mathbb{R}}
\newcommand{\fr}{t}
\newcommand{\pcentroid}{\hat{p}}
\newcommand{\vp}{\mathbf{p}}
\newcommand{\mI}{\mathbf{I}}
\newcommand{\mR}{\mathbf{R}}
\newcommand{\cL}{\mathcal L}
\newcommand{\cU}{\mathcal U}
\newcommand{\cC}{\mathcal C}

\newcommand{\comment}[1]{}
\newcommand{\MSE}[0]{\text{SE}}
\newcommand{\NSE}[0]{\text{NSE}}
\newcommand{\SENL}[0]{\text{SENL}}

\newcommand{\OURM}[0]{\text{\bf $S_\MSE$}}
\newcommand{\OURN}[0]{\text{\bf $S_\NSE$}}
\newcommand{\OURR}[0]{\text{\bf $S_\NSE+M_\NSE$}}
\newcommand{\OURW}[0]{\text{\bf $S_\NSE+M_\NSE + R_\NSE$}}
\newcommand{\OURMSE}[0]{\text{\bf $S_\MSE+M_\MSE + R_\MSE$}}

\newcommand{\argmin}{\operatornamewithlimits{argmin}}

%% file: tex/abstract.tex
\begin{abstract}
Accurate 3D human pose estimation from single images is possible with sophisticated deep-net architectures that have been trained on very large datasets. However, this still leaves open the problem of capturing motions for which no such database exists. 
Manual annotation is tedious, slow, and error-prone. In this paper, we propose to replace most of the annotations by the use of multiple views, at training time {\it only}. Specifically, we train the system to predict the {\it same} pose in all views. Such a consistency constraint is necessary but not sufficient to predict accurate poses. We therefore complement it with a supervised loss aiming to predict the  {\it correct}  pose in a small set of labeled images, and with a regularization term that penalizes drift from initial predictions.
Furthermore, we propose a method to estimate camera pose jointly with human pose, which lets us utilize multi-view footage where calibration is difficult, e.g., for pan-tilt or moving handheld cameras.
We demonstrate the effectiveness of our approach on established benchmarks, as well as on a new Ski dataset with rotating cameras and expert ski motion, for which annotations are truly hard to obtain.
\end{abstract}

%% file: tex/intro.tex
\section{Introduction}

With the advent of Deep Learning,  the effectiveness of monocular 3D human pose estimation algorithms has greatly improved. This is especially true when capturing human motions for which there is enough annotated data to properly train the deep nets. Walking and upright poses are prime examples of this, and state-of-the-art algorithms~\cite{Pavlakos16,Tome17,Popa17,Martinez17,Mehta17b} now deliver impressive real-time results in uncontrolled environments. However, this is not yet the case for more unusual motions for which the training data is harder to obtain, such as sports. Skiing is a good example of this, because pose estimation is crucial to biomechanical and performance analysis, and data cannot easily be acquired in a laboratory. 

\begin{figure}
	\includegraphics[width=\linewidth]{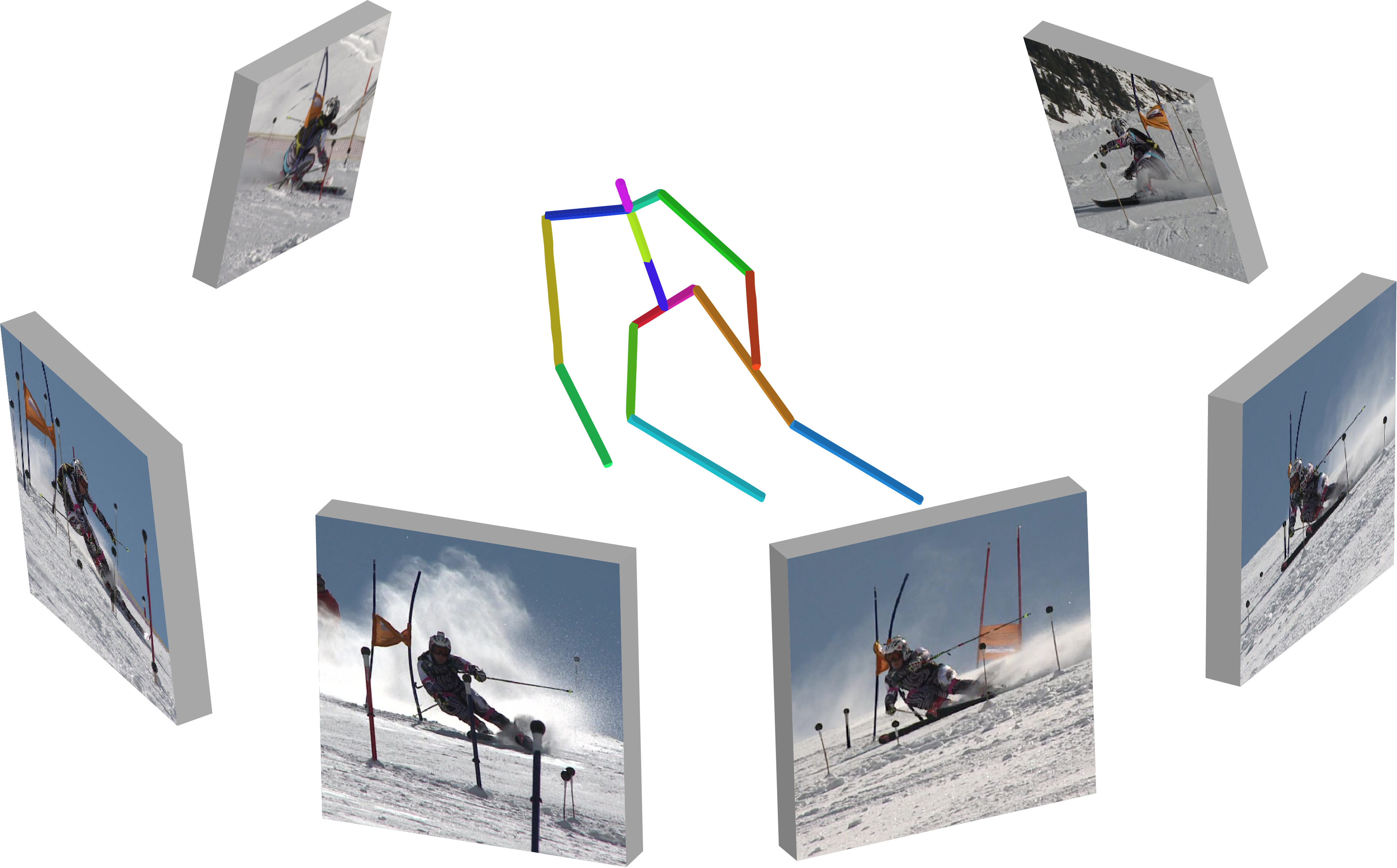}%
	\caption{\label{fig:multi-view-intro}%
		{\bf Multi-view constraints as weak supervision.} Our approach lets us effectively train a deep network to predict 3D pose for actions where only little annotated data is available. This is the case for skiing, which cannot be captured in a laboratory.}
\end{figure}

The brute-force approach to tackle such unusual motions would be to annotate video data. However, achieving high accuracy would require a great deal of annotation, which is tedious, slow, and error-prone. As illustrated by Fig.~\ref{fig:multi-view-intro}, we  therefore propose to replace most of the annotations by the use of multiple views, at training time {\it only}. Specifically, we use them to provide weak supervision and force the system to predict the {\it same} pose in all views. 

While such view consistency constraints increase accuracy, they are unfortunately not sufficient. For example, the network can learn to always predict the same pose, independently of the input image. To prevent this, we use  a small set of images with ground-truth poses, which serve a dual purpose. First, they provide strong supervision during training. 
Second, they let us regularize the multi-view predictions by encouraging them to remain close to the predictions of a network trained with the scarce supervised data only. 

In addition, we propose to use a normalized pose distance to evaluate all losses involving poses.
It disentangles pose from scale, and we found it to be key to maintain accuracy when the annotated data is scarce.

Our experiments demonstrate the effectiveness of our weakly-supervised multi-view training strategy on several datasets, including standard 3D pose estimation benchmarks and competitive alpine skiing scenarios, in which annotated data is genuinely hard to obtain. Not only does our approach drastically cut the need for annotated data, \comment{at no loss in performance \MS{Compared to what?},} but it also increases the robustness of our algorithm to viewpoint and scale changes.  

\comment{
In sum, our contributions are:
\begin{itemize}
	\item A semi-supervised learning approach that utilizes multi-view sequences for weak-supervision.
	\item A robust multi-view loss that utilizes multiple views during NN-training without explicit triangulation.
	\item The design of an adversarial network that counters trivial solutions of the multi-view loss.
\end{itemize}
We demonstrate qualitatively and quantitatively that these contributions constitute a method that is more efficient than existing solutions while requiring a smaller amount of annotated labels. 
}

%% file: tex/related.tex
\section{Related work}
\label{sec:related}

Algorithms for 3D human pose estimation from single images~\cite{Pavlakos16,Tome17,Popa17,Martinez17,Mehta17b,Rogez17,Pavlakos17,Zhou17a,Tekin17a,Sun17}  have now matured to the point where they can operate in real-time and in the wild. They typically rely on sophisticated deep-net architectures that have been trained using very large \comment{training} datasets.  However, dealing with motions for which no such database exists remains an open problem. In this section, we discuss the recent techniques that tackle this aspect.

\vspace{-3mm}
\paragraph{Image Annotation.}An obvious approach is to create the required datasets, which is by no means easy and has become a research topic in itself. 
In a controlled studio environment, marker-suits \cite{Ionescu14b} and marker-less motion capture systems \cite{Mehta17a,Joo15,Pavlakos17} can be used to estimate the pose automatically.  While effective for a subset of human activities, these methods do not generalize well to in-field scenarios in which videos must be annotated manually or using semi-automated tools~\cite{Lassner17}. However, even with such tools, the annotation process remains costly, labor-intensive and error-prone at large scales.

\vspace{-3mm}
\paragraph{Data Augmentation.} An attractive alternative is to augment a small labeled dataset with synthetically generated images. This was done in~\cite{Mehta17a,Rhodin16} by replacing the studio background and human appearance of existing datasets with more natural textures, and in~\cite{Rogez16} by generating diverse images via image mosaicing. Several authors have proposed to leverage recent advances in computer graphics and human rendering~\cite{Loper15} to rely on fully-synthetic images~\cite{Chen16,Varol17}. However, the limited diversity of appearance and motion that such simulation tools can provide, along with their not yet perfect realism, limits both the generality and the accuracy of networks trained using only synthetic images.

\vspace{-3mm}
\paragraph{Weak supervision.} In this paper, this is the approach we focus on by introducing a weakly-supervised multi-view training method. It is related in spirit but different in both task and methodology to the method on geometric supervision of monocular depth estimation from stereo views of Garg \etal~\cite{Garg16}, the multi-view visual hull constraint used for reconstruction of static objects by Yan \etal~\cite{Yan16}\NEW{, and the differentiable ray-potential view-consistency
used by Tulsiani \etal \cite{Tulsiani17}}. Weak supervision has been explored for pose estimation purposes in~\cite{Zhou17a}, which involves complementing fully-annotated data with 2D pose annotations. 
\NEW{Furthermore, Simon \etal \cite{Simon17} iteratively improve a 2D pose detector through view consistency in a massive multi-view studio, using RANSAC and manual intervention to remove outliers.
While effective for imposing reprojection constraints during training, these methods still require extensive manual 2D annotation in the images featuring the target motions and knowledge of the external camera matrix.}
By contrast, the only manual intervention our approach requires is to supply the camera intrinsic parameters, which are either provided by the manufacturer or can be estimated using standard tools. 

%% file: tex/method.tex
\begin{figure}[t]
	\includegraphics[width=\linewidth]{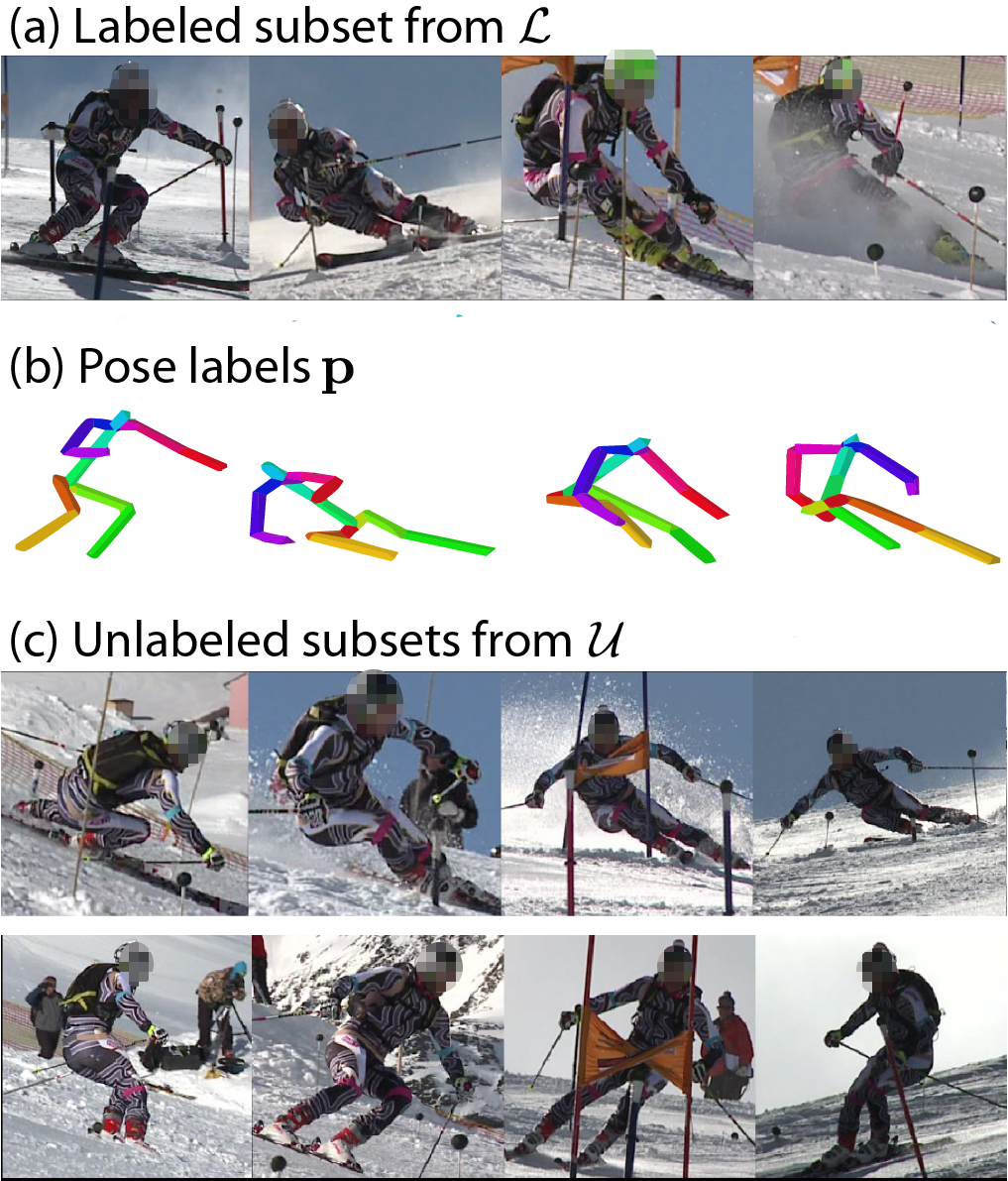}%
	\caption{\label{fig:batchExample}
	{\bf Labeled and unlabeled examples.} The loss $S(\theta,\cL)$ acts on the images (a) with associated poses (b), whereas the view consistency loss $M(\theta,\cU)$ exploits unlabeled images of different skiers that are taken at the same time (c). In (c) each column depicts the same camera and each row the same time $t$.}
\end{figure}

\section{Approach}
\label{sec:approach}
\label{sec:method}
Our goal is to leverage multi-view images, for which the true 3D human pose is unknown, to train a deep network to predict 3D pose from a {\it single} image. To this end, we train our network using a novel loss function that adds view-consistency terms to a standard supervised loss evaluated on a small amount of labeled data and a regularization term that penalizes drift from initial pose predictions.
This formulation enables us to use unlabeled multi-view footage by estimating jointly the body pose in a person-centered coordinate system and the rotation of that coordinate system with respect to the cameras.

Formally, let $f_\theta$ denote the mapping, with parameters $\theta$, encoded by a CNN taking a monocular image $\mI\in\R^{w\times h \times 3}$ as input and producing a 3D human pose $\vp=f_\theta(\mI)\in\R^{3\times N_J}$ as output, where $N_J$ is the number of human joints in our model and the $k^{th}$ column of $\vp$ denotes the position of joint $k$ relative to the pelvis. Furthermore, let $\cL = \{(\mI_i, \vp_i)\}_{i=1}^{N_s}$ be a set of supervised samples, with corresponding images and poses, and $\cU = \{ (\mI_t^j ) _{j=1}^{n_t} \}_{t=1}^{N_u}$ be a set of $N_u$ unsupervised samples, with the $t^{th}$ sample consisting of $n_t$ views of the same person acquired at the same time $t$, as illustrated in Fig.~\ref{fig:batchExample}.
To train our model, we minimize the loss function 
\begin{equation}
L_f(\theta) = M(\theta,\cU) + S(\theta,\cL) + R(\theta,\cU) \; ,
\label{eq:lossF}
\end{equation}
with respect to the network parameters $\theta$, where $M$ is an unsupervised loss that enforces prediction consistency across multiple views, $S$ is a supervised loss that operates on the labeled data, and $R$ is a regularization term.
Below, we describe these three terms in more detail. 

\vspace{-0.3cm}
\paragraph{Multi-View Consistency Loss.}
One of our contributions is to leverage multi-view images as a form of weak supervision for human pose estimation. To this end, our first loss term $M(\cdot)$ encodes view consistency constraints. Specifically, this term penalizes variations other than rigid transformations between the predictions obtained from two or more different views of the same person at the same time. Since our pose vectors are centered at the pelvis, we can ignore camera translation and take rigid transformations to be rotations only. 
We therefore write 
\begin{eqnarray}
M_C(\theta,\cU) & = & \frac{1}{N_u}\sum_{t = 1}^{N_u} \frac{1}{n_t} \sum_{c = 1}^{n_t}  C(\mR_t^c f_\theta(\mI_t^c),\bar{\vp}_t) \; , \label{eq:consistencyLoss}\\
\bar{\vp}_t     & = &  \frac{1}{|\Omega_t|}\sum_{\Omega_t} \mR_t^c f_\theta(\mI_t^{c}) \nonumber \; ,
\end{eqnarray}
where $\mR^c_t$ denotes the rotation matrix for camera $c$ and sample $t$, $\bar{\vp}_t$ is a reference pose for sample $t$, computed as a robust mean of the individual poses obtained by averaging over the consensus set $\Omega_t$ of views whose predictions agree. $\Omega_t$ is \ms{obtained}
 with a deterministic variant of the traditional RANSAC~\cite{Fischler81}, as the subset of $f_\theta(\mI_{t}^{c})^{n_t}_{c=1}$ with the largest agreement in mean pose. $C(\cdot)$ denotes the distance between poses. 

\begin{figure}
	\includegraphics[width=\linewidth,trim={0 0 0 0},clip]{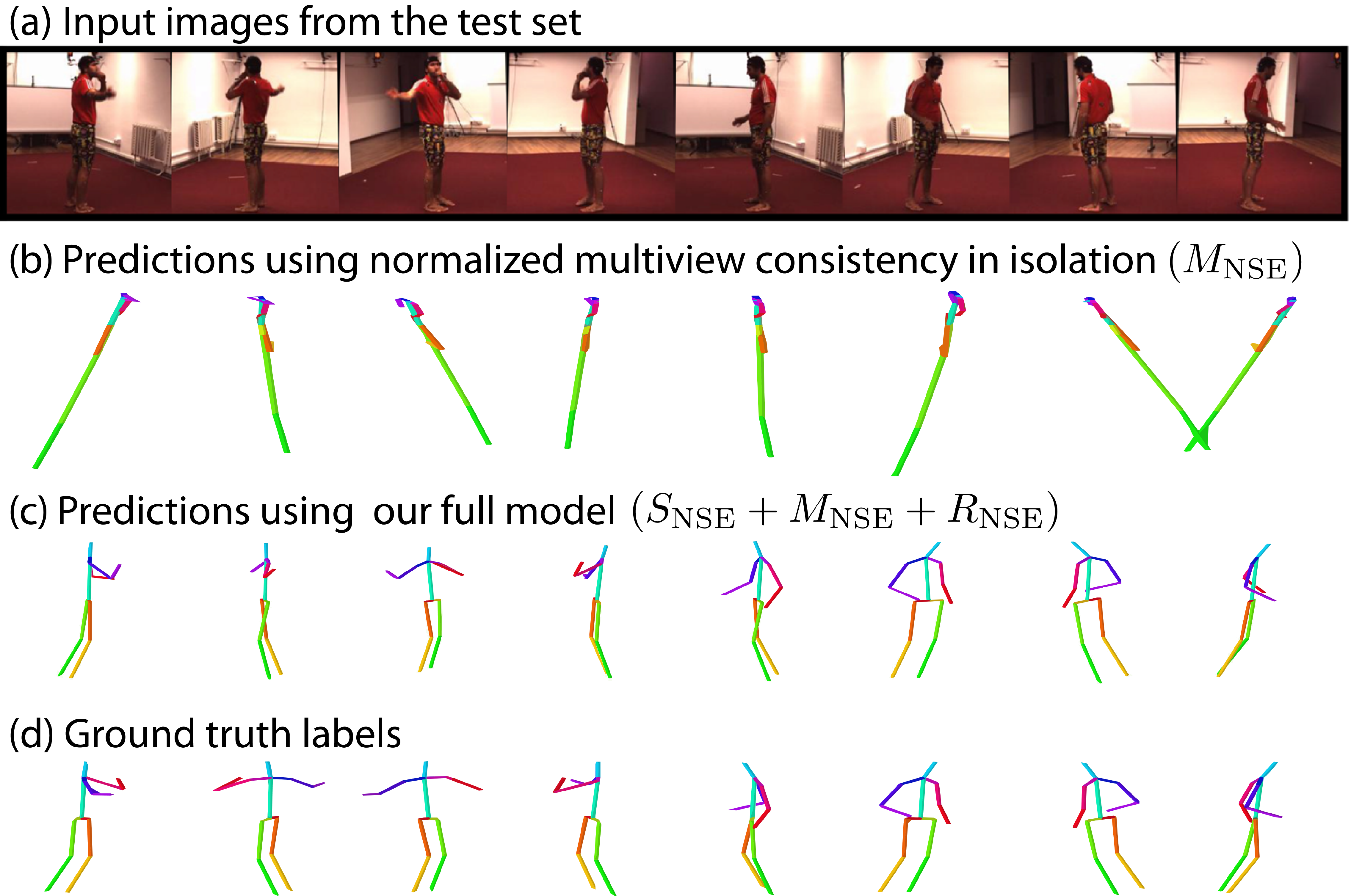}%
	\caption{\label{fig:squeezing}%
		{\bf The multi-view consistency constraint} evaluated on test images (a).
		Without the supervised term $S_\NSE$ and regularization term $R_\NSE$ (b) predictions are biased towards elongates poses. Only the full model (c) can exploit the unlabeled examples and predict correct poses.
	}
\end{figure}

\vspace{-0.3cm}
\subparagraph{Distance between poses.} We could take $C(\cdot)$ in Eq.~\ref{eq:consistencyLoss} to be the squared error
\begin{equation}
\MSE{}(\vp_1, \vp_2) = \| \vp_1 - \vp_2 \|^2 \; ,
\label{eq:mse}
\end{equation}
where $\| \vp \|$ is the vector norm of $\vp$.
However, scale is difficult to estimate from single images. Furthermore, $\MSE(\cdot)$ can be trivially minimized by taking $\vp_1=\vp_2=0$. 
\comment{As shown in Fig.~\ref{fig:squeezing},}
This can lead to undesirable behaviors when neither $\vp_1$ nor $\vp_2$ is fixed, which is the case when working with the proposed unsupervised term.\comment{unsupervised data only.} 
\comment{will be the case in the unsupervised case discussed below. \PF{Does Fig.~\ref{fig:squeezing} really illustrate the $\vp_1=\vp_2=0$ case of something else?} \MS{I agree that this example in Fig.~\ref{fig:squeezing} is not convincing at all. The poses look much better with SE than with NSE.} }
We therefore introduce a new distance that is normalized for scale and expressed as
\begin{equation}
\NSE{}(\vp_1, \vp_2) = \left|\left| \frac{\vp_1}{\|\vp_1\|} - \frac{\vp_2}{\|\vp_2\|}\right|\right|^2 \; .
\label{eq:nmse}
\end{equation}
\NEW{It has a similar influence as enforcing constant bone length with a geometric constraint.}
We will see in the results section that using $\NSE$ instead of $\MSE$ during training substantially increases accuracy, especially  when there are only very few labeled samples.

\vspace{-0.3cm}
\subparagraph{Estimating the rotations.} 

For static capture setups, the rotation $\mR^c_t$ is constant for each camera $c$ across all times $t$ and can easily be estimated using standard tools. However, in the wild and when the camera is moving, estimating the changing $\mR^c_t$s becomes more difficult.  In line with our goal of minimizing the required amount of manual intervention, we use the subjects and their estimated poses as calibration targets. 

Without loss of generality, let $\mR_t^{1}$ be the identity, meaning that all other rotations are expressed with respect to the coordinate system of the first camera. We estimate them as 
\begin{equation}
\mR_t^c = \argmin_\mR \left\Vert \frac{\mR f_\theta(\mI_t^{c})}{\|f_\theta(\mI_t^{c})\|} - \frac{f_\theta(\mI_t^{1})}{\|f_\theta(\mI_t^{1})\|} \right\Vert^2\; ,
\label{eq:procrustes}
\end{equation}
where the $f_\theta(\mI_t^{c})$ are the current pose estimates. This corresponds to the rotational part of Procrustes analysis \cite{Horn87}. \NEW{For the sake of robustness, it is only computed on the torso joints, not the limb joints.}

During training, we do not backpropagate the gradient through this rotation matrix computation, which would require a singular value decomposition and is intricate to differentiate~\cite{Ionescu15}. Instead, we simply fix the rotations to update the network parameters $\theta$ and update $\mR_t^c$ after each gradient iteration.
This makes our approach independent from the camera rotation. It is also independent from the camera position because it predicts hip-centered poses, as is common in the field. Therefore, unlike other recent multi-view approaches~\cite{Simon17,Pavlakos17}, we do {\it not} require a full camera calibration.
We only need the intrinsics.  

\vspace{-0.3cm}
\paragraph{Supervised Regression Loss.} 

As illustrated by Fig.~\ref{fig:squeezing}, only using the multi-view consistency loss would yield trivial solutions where all predicted poses would be random but identical. To prevent this, we propose to also use a small amount of supervised data. To this end, we define a loss that penalizes deviations between the predicted and ground-truth poses. We write it as
\begin{equation}
S_C(\theta,\cL) = \frac{1}{N_s}\sum_{i=1}^{N_s} C(f_\theta(\mI_i), \vp_i) \; ,
\label{eq:supervisedLoss}
\end{equation}
where $C$ is one of the distances introduced before, that is, it can be either the squared error of Eq.~\ref{eq:mse} or the scale-invariant version of Eq.~\ref{eq:nmse}.

\vspace{-0.3cm}
\paragraph{Regularization Loss.}

As shown in Fig.~\ref{fig:modelComponentsValidation}, using both $M(\cdot)$ and $S(\cdot)$ during training improves validation accuracy in the early training stages, thus showing the benefits of our multi-view weak supervision. However, after several epochs, the resulting model overfits, as indicated by the increasing error in Fig.~\ref{fig:modelComponentsValidation}. 
On closer examination, it appeared that the network learns to distinguish the labeled examples from the unlabeled ones and to, again, predict consistent but wrong poses for the latter, as depicted by Fig.~\ref{fig:squeezing} (b).

To prevent this, we introduce an additional regularization term $R_C(\theta, \cU)$, that penalizes predictions that drift too far away from the initial ones during training. To obtain these initial predictions, we use our scarce annotated data to begin training a network using only the loss $S$ and stop early to avoid overfitting, which corresponds to the 0 on the $x$-axis of Fig.~\ref{fig:modelComponentsValidation}. 
Let $\gamma$ be the parameters of this network. Assuming that the labeled poses are representative, the predictions $f_\gamma(\mI_t^c)$ for the unlabeled images will look realistic. We therefore write our regularizer as
\begin{equation}
R_C(\theta, \cU) =\frac{1}{N_u}\sum_{t = 1}^{N_u} \frac{1}{n_t} \sum_{c = 1}^{n_t} C(f_\theta(\mI_t^c), f_\gamma(\mI_t^c)) \; .
\label{eq:regularizationLoss}
\end{equation}
In other words, we penalize 3D poses that drift too far away from these initial estimates. In principle, we could progressively update $\gamma$ during training. However, in practice, we did not observe any improvement and we keep it fixed in all experiments reported in this paper. 

Our complete approach is summarized in Algorithm~\ref{alg:method}.

\SetAlgoNoLine
\begin{algorithm}[t]
	\small
	\DontPrintSemicolon
	\KwData{Labeled training set $\cL$ and unlabeled set $\cU$}
	\KwResult{Optimized neural network parameters $\theta$}
	Pre-training of $f_\theta$ on $\cL$ through $S_\NSE$\;
	$\gamma \gets \theta$\;
	\For{\# \emph{of gradient iterations}}{
		Select a random subset of 8 examples from $\cL$\;
		Select 8 examples from $\cU$ such that the first four ($n_t=4$) as well as the last four are taken at the same time $t$\;
		\If{$\mR$ \emph{not available}}{Estimate rotations $\mR_t^c$ for each quadruple}
		Infer consensus sets $\Omega_t$\;
		Compute reference poses $\bar{\vp}_t$\;
		Optimize $L_f(\theta)$ with respect to $\theta$ using Adam
	}
	\caption{\label{alg:method}
			Summary of our weakly supervised training method using the default parameters}
\end{algorithm}

\paragraph{Implementation Details.}

We rely on the ResNet-50 architecture~\cite{He16} up to level 5. In order to regress the pose vector output from convolutional features, we replace level 5 with two convolutional layers and a fully-connected layer. The first convolutional layer we added has a $3\times3$ kernel size and 512 output channels. The second one has a $5\times5$ kernel size and 128 output channels. The first three levels are initialized through transfer learning by pre-training on a 2D pose estimation task, as proposed by \cite{Mehta17a}, and then kept constant. For the weak supervision experiments, the network is pre-trained on $\cL$.

During training, we use mini-batches of size 16. Each one contains 8 labeled and 8 unlabeled samples. A consensus set size $|\Omega|$ of two was most effective in our examples. If more than four camera views are available, a random subset of cardinality four is chosen.
Since the full objective function $L_f$ of Eq.~\ref{eq:lossF} is the sum of three terms, their respective influence has to be adjusted. We have found empirically that weighting the supervised loss $S$ of Eq.~\ref{eq:supervisedLoss} and regularizer $R$ of Eq.~\ref{eq:regularizationLoss} by 100 and the unsupervised loss $M$ of Eq.~\ref{eq:consistencyLoss} by 1 worked well for all experiments. We used the Adam optimizer with a constant learning rate of $0.001$. 
All examples in our training database are of resolution $256\times256$, augmented by random rotations of $\pm20^{\circ}$  and scaled by a factor between 0.85 and 1.15. Furthermore, we correct for the perspective effect
as described in \cite{Mehta17a}, which requires the camera intrinsics.

%% file: tex/evaluation.tex
\section{Results}
\label{sec:results}

In the following, we quantify the improvement of our weak-supervision approach and evaluate its applicability in diverse situations. We will provide videos and further qualitative results in the supplementary material.

\vspace{-3mm}
\paragraph{Datasets.}
We first test our approach on the well-known Human3.6M (H36M)~\cite{Ionescu14b} dataset to compare it to other state-of-the-art methods and on the more recent MPII-INF-3DHP (3DHP) set~\cite{Mehta17a} to demonstrate how it generalizes to  different viewpoints and outdoor scenes. In both cases, the images were recorded using a calibrated multi-camera setup, which makes these dataset easily exploitable for us.

\begin{figure}
	\centering
\includegraphics[width=0.25\linewidth,height=0.25\linewidth,trim={3px 3px 3px 3px},clip]{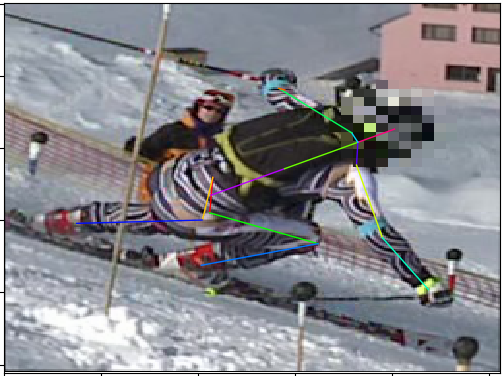}%
\includegraphics[width=0.25\linewidth,height=0.25\linewidth,trim={3px 3px 3px 3px},clip]{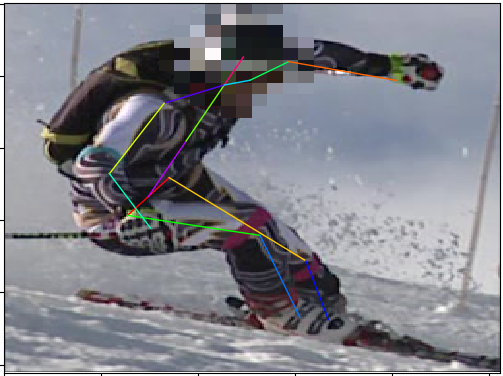}%
\includegraphics[width=0.25\linewidth,height=0.25\linewidth,trim={3px 3px 3px 3px},clip]{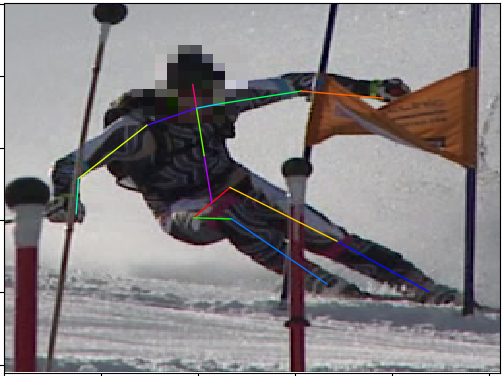}%
\includegraphics[width=0.25\linewidth,height=0.25\linewidth,trim={3px 3px 3px 3px},clip]{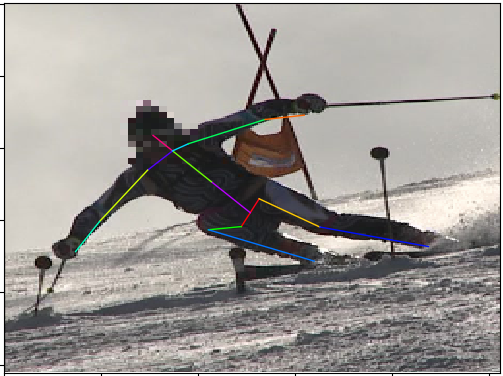}%
	\caption{\label{fig:ski_annotation}
		\NEW{{\bf Alpine ski dataset.} The dataset provides six camera views as well as corresponding 3D pose. Four example views are shown. The 3D pose matches accurately when reprojected onto the input views, here shown by the stick-figure overlay.}}
\end{figure}

To highlight the effectiveness of our approach when using moving cameras to capture specialized motions that cannot be performed indoors, we also introduce a multi-view ski dataset that features competitive racers going down alpine slalom courses. The setup is shown in Fig.~\ref{fig:ski_overview}. \NEW{Details on the capture and annotation process are explained in the supplemental document. Re-projection of the 3D annotation shows a very accurate overlap with the images, see Fig.~\ref{fig:ski_annotation}.} We make it available to facilitate future work towards reliable monocular pose reconstruction with weak supervision (\href{http://cvlab.epfl.ch/Ski-PosePTZ-Dataset}{cvlab.epfl.ch/Ski-PosePTZ-Dataset}).

\begin{figure}
	\centering
	\includegraphics[height=0.35\linewidth,trim={0 0 0 0},clip]{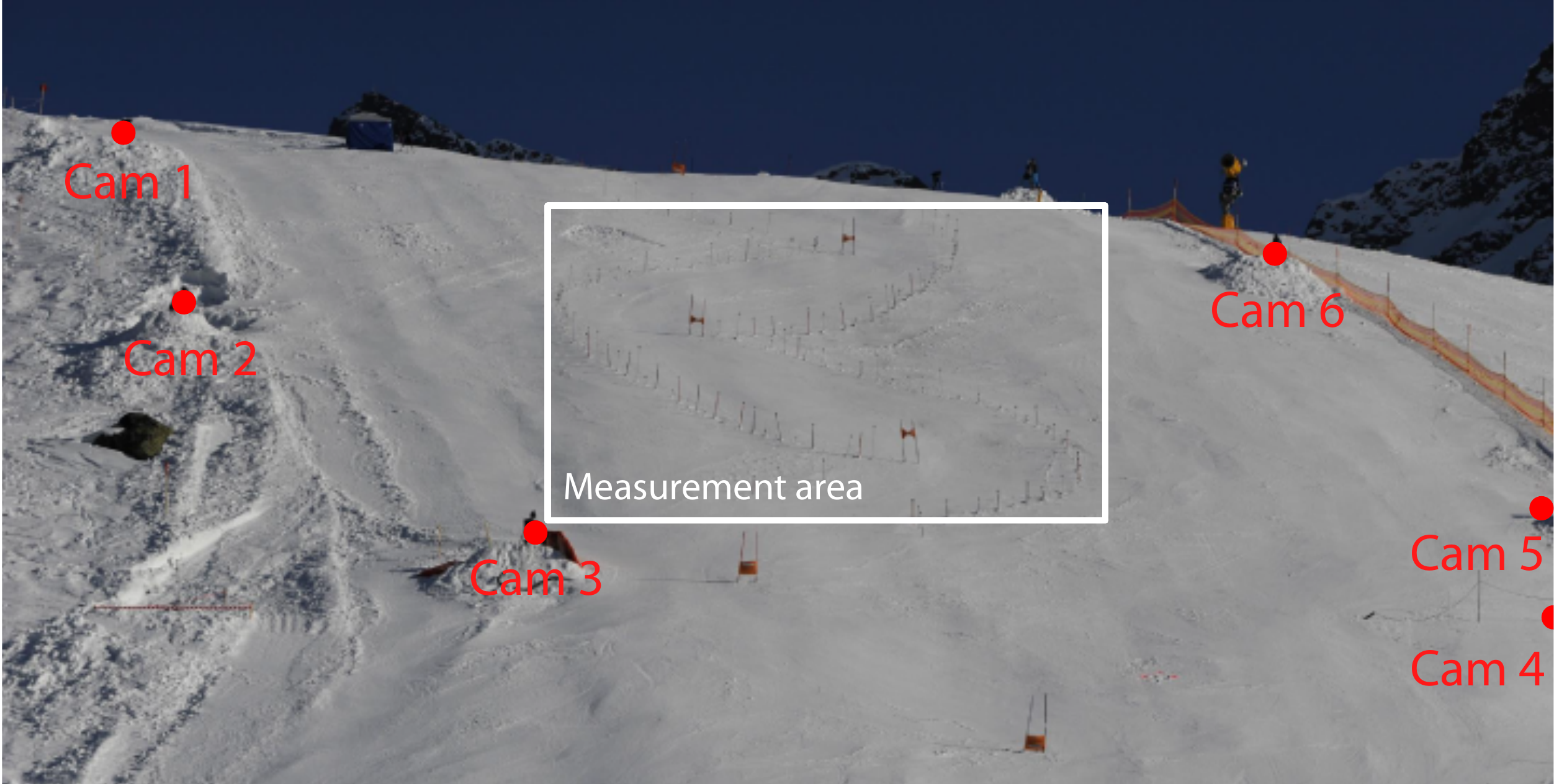}
	\,
	\includegraphics[height=0.35\linewidth,trim={0 0 0 0},clip]{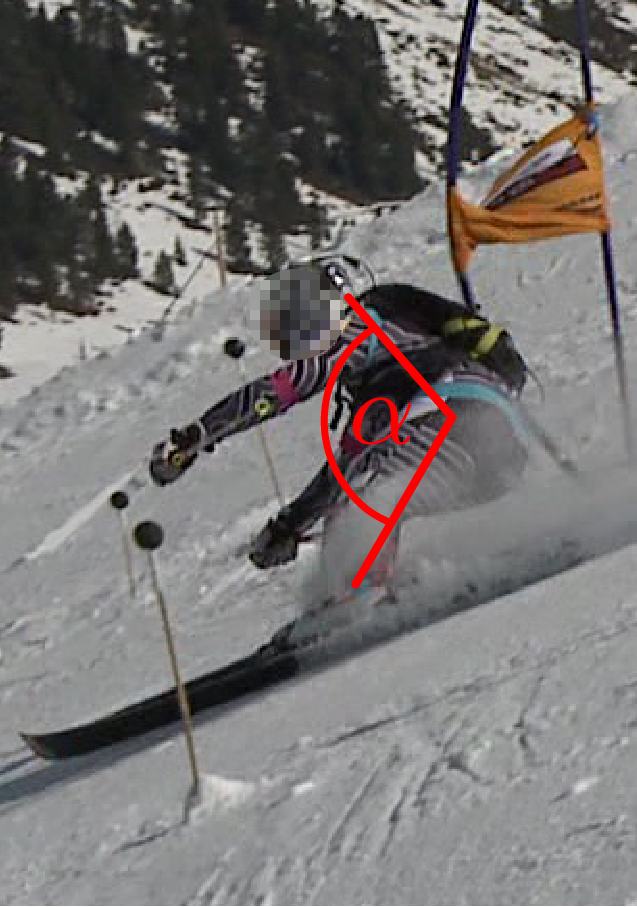}
	\caption{\label{fig:ski_overview}
		{{\bf The alpine ski measurement setup.} (Left) Six pan-tilt-zoom cameras are placed in a rough circle around the race course. (Right) Metrics such as hip flexion are commonly used for performance analysis.}}
\end{figure}

\vspace{-3mm}
\paragraph{Metrics.}

\input{fig/h36m_table}

\input{fig/h36m_table_loss_function}
We evaluate pose accuracy in terms of the mean per joint position error (MPJPE) and the percentage of correct keypoints (PCK), that is, the percentage of joints whose estimated position is within 15cm of the correct one. To make both measures independent from the subjects' height, we systematically apply a single scale factor to the prediction so as to minimize the squared distance $\MSE$ between label and prediction. We refer to the resulting normalized metrics as NMPJPE and NPCK. Since orientation is left unchanged, this is a less constraining transformation than the more commonly used Procrustes alignment, to which we refer as PMPJPE. 
For skiing, we also compute four specialized metrics commonly used in this scenario \cite{Fasel18,Fasel16}. 
\NEW{The first two are the COM-hip joint distance and the COM-ankle distance being representative of relative COM position.
The COM is computed from 3D pose according to the average body weight distribution~\cite{Clauser71}.}
The other two are the hip-flexion and knee-flexion angles. These metrics have been extensively used in sking related research before and are depicted by Fig.~\ref{fig:ski_overview}~(right). 

\paragraph{Baseline.}
In the top portion of Table~\ref{tb:H36M}, we report the MPJPE and  NMPJPE values of fully-supervised methods on H36M using the same protocol as in~\cite{Li14a,Zhou17a,Mehta17a,Tekin17a,Pavlakos16,Pavlakos17}: Five subjects (S1, S5, S6, S7, S8) are used for training and the remaining two (S9, S11) for testing. The 3D position of the 16 major human joints is predicted relative to the pelvis. The training and test sets are subsampled at 10fps, to reduce redundancy and validation time. 

It shows that our modified ResNet-50 architecture, introduced in Section~\ref{sec:approach}, yields results that are representative of the state-of-the-art methods, despite being considerably simpler---and also faster to train and experiment with. For example,
the approach of~\cite{Mehta17a} uses a residual network with twice as many layers as ours, along with a complex learning rate scheduling. The popular stacked-hourglass network used by some of the most recent methods~\cite{Martinez17,Zhou17a,Tekin17a} is even more complex, which results in long training and testing times \NEW{of 32 ms per batch}. %
\NEW{The volumetric network of Pavlakos \etal~\cite{Pavlakos16} even takes 67~ms.}
Popa \etal~\cite{Popa17} use a complex iterative multitask architecture and incorporate semantic segmentation, that is, labels that are not used by any of the other methods. This justifies choosing our modified ResNet-50 architecture, \NEW{which has a runtime of only 11 ms}.

\NEW{For training the baseline, we test the \NSE{} and \MSE{} loss using the dataset mean and standard deviation normalization as proposed by \cite{Sun17}. The \NSE{} loss performs better than \MSE{} when the labeled training set is small and in combination with the multi-view constraint, see Table~\ref{tb:H36M_loss_function}.}
We will refer to it as \OURN{} in the remainder of this section. Since our approach does not depend on the architecture of the base classifier, the slightly higher accuracies obtained by more complex ones would translate to our weakly-supervised results if we were to use them as our model.

\subsection{Multi-View Supervision on Human3.6M}

\begin{figure}
	\centering
	\includegraphics[width=1\linewidth,trim={0 0 0 0},clip]{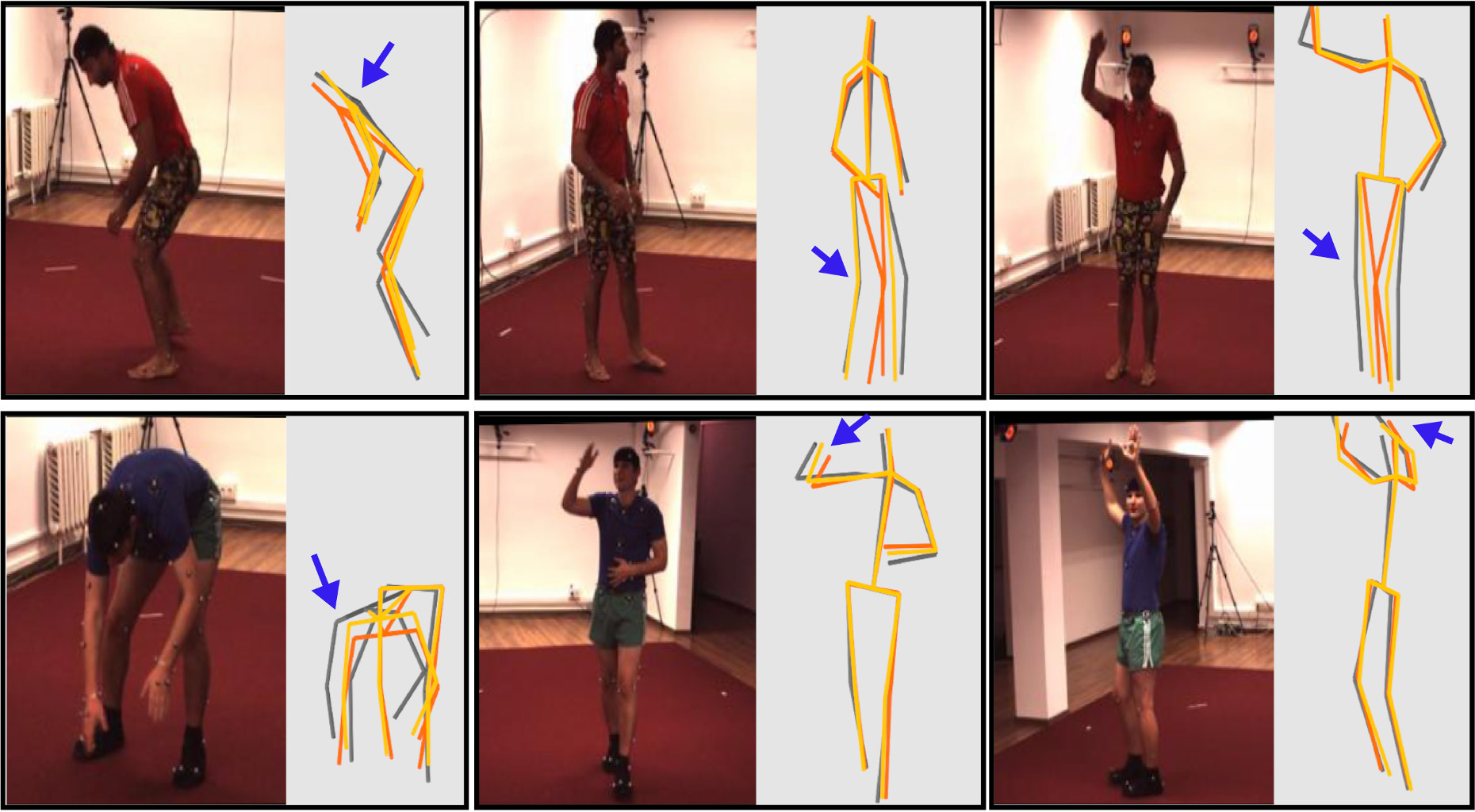}%
	\caption{\label{fig:h36mQualitative}%
		\OURW{} (yellow) and \OURN{} (orange) results overlaid on the ground truth label (black). Out of 2800 test images, the NMPJE is improved by more than 10 mm  in 813 of them and degraded in only 259. Crouching and extreme stretching motions are improved most often.}
\end{figure}

To test our multi-view weak supervision scheme, we split the standard training set into a supervised set $\cL$ and an unsupervised one $\cU$. In $\cL$, we use the images and associated ground-truth poses, while in $\cU$, we only use the camera orientation and bounding box crop information. In the bottom portion of Table~\ref{tb:H36M}, we report our results for two different splits: Either two subjects, S1 and S5, in $\cL$ and the other three in $\cU$, or only S1 in $\cL$ and the other four in $\cU$. The numbers we report are averages for all actions and we supply a detailed breakdown in the supplementary material. In both cases, the performance of \OURN{} trained only on  $\cL$ is lower than when using the full training set. Importantly, our approach ($\OURW{}$) allows us to recover 5 mm in NMPJPE. That is, 28\% and 42\% of the difference to the full set is recovered, thus showing the benefits of our weak supervision.
We show qualitative results in Fig.~\ref{fig:h36mQualitative}.
In all these experiments, we used the known camera rotations. We will discuss what happens when we try to also recover the rotations below.

Fig.~\ref{fig:h36m_graph} summarizes these results. We plot the NMPJPE values as a function of the number of people in $\cL$. As expected, the relative improvement brought about by our weak supervision is larger when we have fewer people in $\cL$, and the order of the methods remains consistent throughout, with \OURW{} being best, followed by \OURR{} and \OURN{}. The behavior is exactly the same when using NPCK instead of NMPJPE as error metric.

To demonstrate the importance of the regularization term of Eq.~\ref{eq:regularizationLoss}, we also trained our network using a loss function that is the sum of the multi-view loss of Eq.~\ref{eq:consistencyLoss} and the supervised loss of Eq.~\ref{eq:supervisedLoss}, but without regularization. We refer to this as \OURR{}, which we found much more difficult to train. As shown in Fig.~\ref{fig:modelComponentsValidation}, performing too many training iterations decreases accuracy, and we therefore had to resort to early stopping. Our complete model, with the additional regularization term, does not suffer from this issue.

\paragraph{Network initialization}
\NEW{Initializing the weights with ImageNet instead of pre-training on 2D human pose yields an error of 95.4 mm NMPJPE for our fully-supervised baseline. Using only S1 and S5 for supervision yields 159.0 mm.
Using the unsupervised subjects S6-S8 improves by 16.1 mm, or 25.3 \% of the gap. This is a larger absolute but slightly smaller relative improvement compared to the results obtained by pre-training on the MPII 2D pose dataset. 
For S1 our method improves from $161.4$ to $153.3$ and for $S1+S5+S6$ from $122.9$ to $114.5$.
This shows that our multi-view approach can be applied as is, even in the absence of 2D annotations.}

\input{fig/h36m2}

\input{fig/h36m1}

\input{fig/MPII3DHP}

\subsection{Viewpoint Changes --- MPII-3DHP}

While the prediction accuracy of state-of-the-art methods on Human3.6M saturates at about 60mm, the recent MPII-3DHP dataset targets more challenging motions, such as gymnastics. \comment{with close floor contact.} 
Ground-truth joint positions were obtained via marker-less motion capture from 12 synchronized cameras, which allows the actors to wear more diverse clothing.
We rely on the standard protocol: The five chest-hight cameras and the provided 17 joint \emph{universal} skeletons are used for training and testing. Note that here we compare the methods in terms of the PCK and NPCK metrics, which are more widely used on this dataset and for which a higher value is better.
Table~\ref{tb:3DHP} shows that our baseline \OURN{} is on par with existing methods when trained in a fully supervised manner either on H36M or 3DHP and tested on 3DHP. In the weakly-supervised case, with a single labeled subject, $\OURW{}$ improves  NPCK, as it did in the H36M experiments.

An interesting aspect of 3DHP is the availability of top-down and bottom-up views, which are missing from other pose datasets. This allows us to test the robustness of our approach to new camera views, as shown in Fig.~\ref{fig:3DHP_improvements}. To this end, we train \OURW  with the standard views of S1 as strong supervision and with unlabeled examples from all the views---both standard and not---for S2 to S7. As shown in the bottom portion of Table~\ref{tb:3DHP}, this greatly improves the predictions for the novel views of S8 compared to using the supervised data only. %

\input{fig/MPII3DHP_table}

\subsection{Outdoor Capture--- Competitive Ski Dataset}
\label{sec:ski}

In competitive skiing, the speeds are such that a static camera setup would cover a capture volume too small to acquire more than fractions of a turn. Therefore, most biomechanical studies rely on pan-tilt-zoom (PTZ) cameras to follow the racers over multiple turns, and their pose is estimated by manually annotating the images. The technique we propose here has the potential to eliminate the need for most of these annotations.  

For our experiments, we have access to a training database that was used in an earlier \NEW{methodological} study~\cite{Fasel16}. Six pan-tilt-zoom cameras were used to capture six subjects during two runs each down a \NEW{giant} slalom course. See Fig.~\ref{fig:ski_overview} for an overview. After manual annotation, followed by processing and filtering inaccurate poses, 10k temporal frames remained, which were split into 8481 for training (Subjects 1–5) and 1760 for testing (Subject 6).  

The intrinsic and extrinsic camera parameters were estimated using geodetically measured reference points. In other words, the rotation $\mR_t^c$ of camera $c$ with respect to the reference frame is precisely known for all cameras at all times. To test the ability of our approach to operate without this knowledge, we report results both when using these known rotations and when estimating them by solving the minimization problem of Eq.~\ref{eq:procrustes}.

\input{fig/ski_spoerri}

\vspace{-3mm}
\paragraph{Known rotations}

In Fig.~\ref{fig:evalSKISPOERRI}, as before, we report  our results as a function of the number of skiers used to form the supervised set $\cL$, with the videos of the other ones being used to enforce the multi-view consistency constraints. We express our results in terms of NMPJPE, COM, hip-flexion, and knee-flexion angles. Altogether, \OURW{} systematically improves over \OURN{}, with the improvement monotonically decreasing the more labeled data we use. 
In particular occlusions and extreme poses are improved, as depicted by Fig.~\ref{fig:ski_spoerri_improvements}.

\vspace{-3mm}
\paragraph{Estimated rotations}

The results shown above were computed assuming the rotations to be known, which was the case because the images were acquired using a very elaborate setup. To test the viability of our approach in a less constrained \comment{more unconstrained} setup, such as one  where the images are acquired using hand-held cameras, we repeated the above experiments without assuming the rotations to be known. The results are shown in Fig.~\ref{fig:evalSKIMeasurements} as purple bars.
The improvement brought about by the weak supervision drops with respect to what it was when the rotations were given---for example from 7.4 to 3.4 mm in NMPJPE terms when using the motions of a single skier to form $\cL$---but remains consistent. A similar improvement of 3-5 mm in NMPJPE is maintained on H36M and 3DHP, see the row below the dashed line in Tables~\ref{tb:H36M} and \ref{tb:3DHP}. 

\vspace{-3mm}
\paragraph{Convergence.}

\begin{figure}
	\centering
	\includegraphics[width=0.8\linewidth,trim={0 0 0 3.3cm},clip]{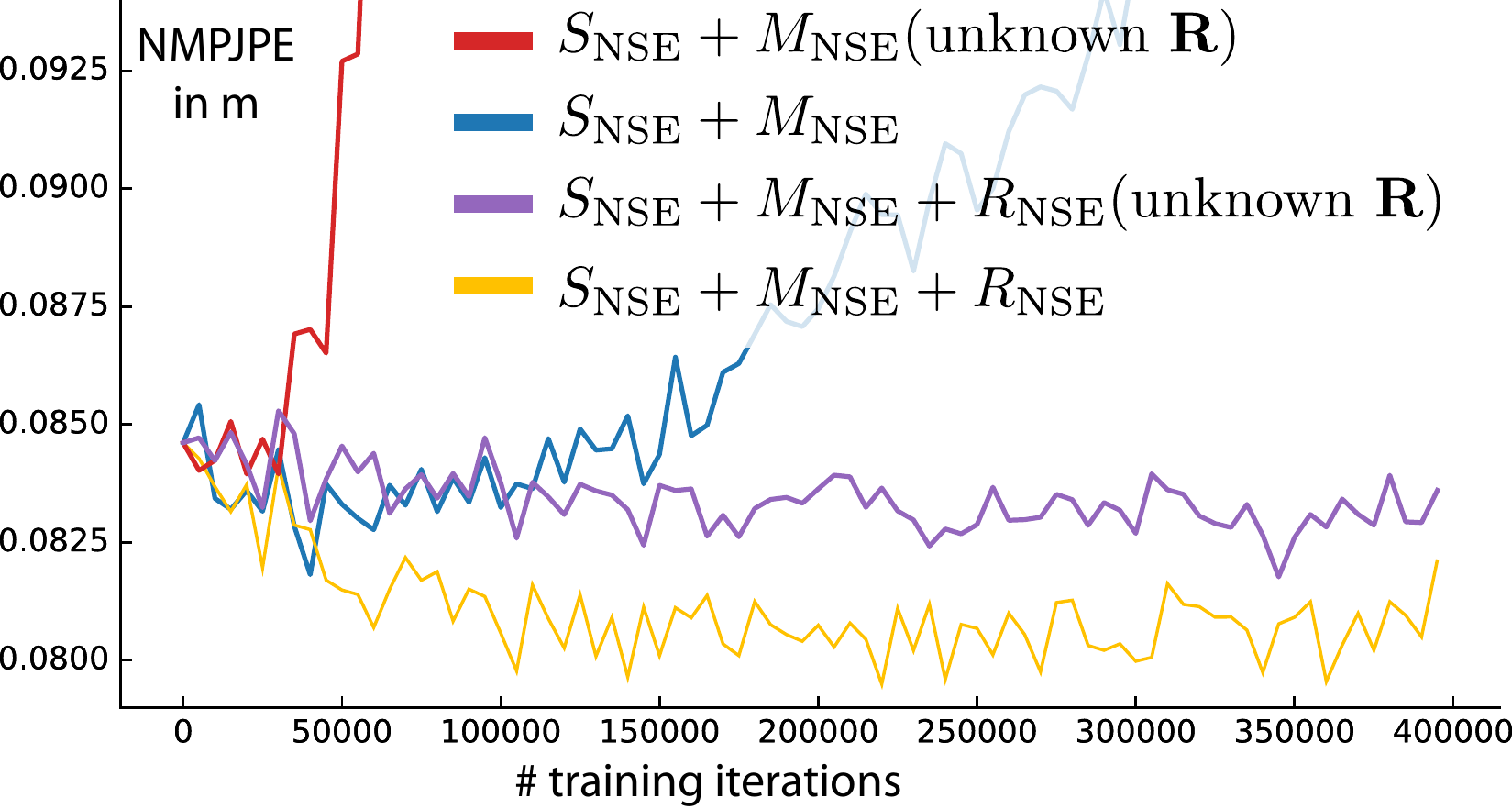}%
	\caption{\label{fig:modelProcrustes}%
		Training error on H36M, with and without estimating the camera rotation $\mR$ alongside the human pose $\vp$. Training only converges with the proposed regularization term.
	}
\end{figure}

In Fig.~\ref{fig:modelProcrustes}, we plot the training errors for the different versions of our approach, including \OURR{}, the variant in which we minimize the multi-view consistency constraint but not the regularization term, for both known and estimated rotations. When regularizing, the loss remains well-behaved and attains a lower minimum in both cases, whereas it diverges when the regularizer is omitted, which confirms its importance.

%% file: fig/h36m_table.tex
\begin{table}
	\center
\resizebox{1\columnwidth}{!}{
		\begin{tabular}{ |c|l|c|c|c| }
			\multicolumn{5}{c}{Supervised training on all subjects of H36M}\\
			\hline
			Training Sub. &Method & MPJPE in mm & NMPJPE in mm & PMPJPE in mm\\
			\hline
			\multirow{9}{*}{\rotatebox[origin=c]{90}{Subjects S1+S5+S6+S7+S8}}
		    &Pavlakos \etal \cite{Pavlakos17} & 118.4 & not available & not available \\
			&Rogez \etal \cite{Rogez17} & {87.7} & not available & 71.6\\
		    &Pavlakos \etal \cite{Pavlakos16} & 71.9 & not available & 51.9\\
			&Zhou \etal \cite{Zhou17a} & 64.9 & not available & not available\\ 
			&Mehta \etal \cite{Mehta17a} &  74.1 & 68.6 & 54.6 \\
			&Tekin \etal \cite{Tekin17a} & {69.7} & not available & {50.1} \\
			&Popa \etal \cite{Popa17}$^{\star}$ & {63.4} & not available & not available\\
			&Martinez \etal \cite{Martinez17} &  \bf{62.9} & not available & \bf{47.7} \\
			&\OURM{} & \NEW{66.8} & \NEW{\bf{63.3}} & \NEW{{51.6}} \\
			&\OURN{}& not applicable & {64.2} & {53.0}\\
			\hline
\multicolumn{5}{c}{}\\
\multicolumn{5}{c}{Weakly-supervised training with varying training sets}\\
			\hline%
			Training Sub. &Method & MPJPE in mm & NMPJPE in mm & PMPJPE in mm\\
			\hline%
			\multirow{2}{*}{\small S1 only, known $\mR$}
			&\OURN{} & not applicable & {83.4} & 68.4\\
			&\OURW{} & not applicable & \bf{78.2} & \bf{64.6}\\
			\hdashline
			\small S1,  unknown $\mR$ &\OURW{}  & not applicable & \bf{80.1} & \bf{65.1} \\
			\hline
			\multirow{2}{*}{\small S1+S5, known $\mR$}
			&\OURN{} & not applicable & {76.1} & 61.8\\
			
			&\OURW{}  & not applicable & \bf{70.8} & \bf{58.5} \\
			\hline
		\end{tabular}
	}
 \vspace{1pt}
	\caption{{\bf H36M results.} (Top portion) In the fully supervised case, our comparatively simple network architecture is competitive when used in conjunction with the $\NSE$ distance function (\OURN{}), but its performance is worse when used in conjunction with $\MSE$ (\OURM{}). Note that the method of~\cite{Popa17} ($^{\star}$) uses silhouette information for training, which none of the other methods do.
		 (Middle portion) Using a single subject to provide the supervised training set and enforcing the consistency constraints on the others (\OURW{}) or not (\OURN{}). (Bottom portion)  Using two subjects  to provide the supervised training set. In both cases, imposing the consistency constraints boosts the accuracy.}
	\label{tb:H36M}
\end{table}

%% file: fig/h36m_table_loss_function.tex
\begin{table}
	\center
\resizebox{1\columnwidth}{!}{
		\begin{tabular}{ |c|l|c|c|c| }
\multicolumn{5}{c}{\NEW{Weakly-supervised training with varying loss functions}}\\
			\hline%
			Training Sub. &Method & MPJPE in mm & NMPJPE in mm \\
			\hline%
			\multirow{4}{*}{\small S1 only, known $\mR$}
			&\OURM{} & 99.6 & 91.5\\
			&\OURMSE{} & \bf{98.5} & {88.8}\\
			&\OURN{} & not applicable & {83.4}\\
			&\OURW{} & not applicable & \bf{78.2}\\
			\hline
			\multirow{4}{*}{\small S1+S5, known $\mR$}
			&\OURM{} & 90.3 & 77.5\\
			&\OURMSE{} & \bf{89.0} & {74.7}\\
			&\OURN{} & not applicable & {76.1}\\		
			&\OURW{}  & not applicable & \bf{70.8}\\
			\hline
		\end{tabular}
	}
 \vspace{1pt}
	\caption{{\bf Different pose distances on H36M.} \NEW{Our multi-view constraint improves reconstructions, both with the commonly used $\MSE$ distance (\OURMSE{}) and the normalized $\NSE$ version (\OURW{}). However, reconstruction accuracy is higher and improvements are larger for the proposed normalized loss (compare second and fourth rows). This improvement is consistent across varied training sets (top vs.~bottom portion).}}
	\label{tb:H36M_loss_function}
\end{table}

%% file: fig/h36m2.tex
\begin{figure}
	\centering
	\includegraphics[width=0.8\linewidth,trim={0 0 0 3.6cm},clip]{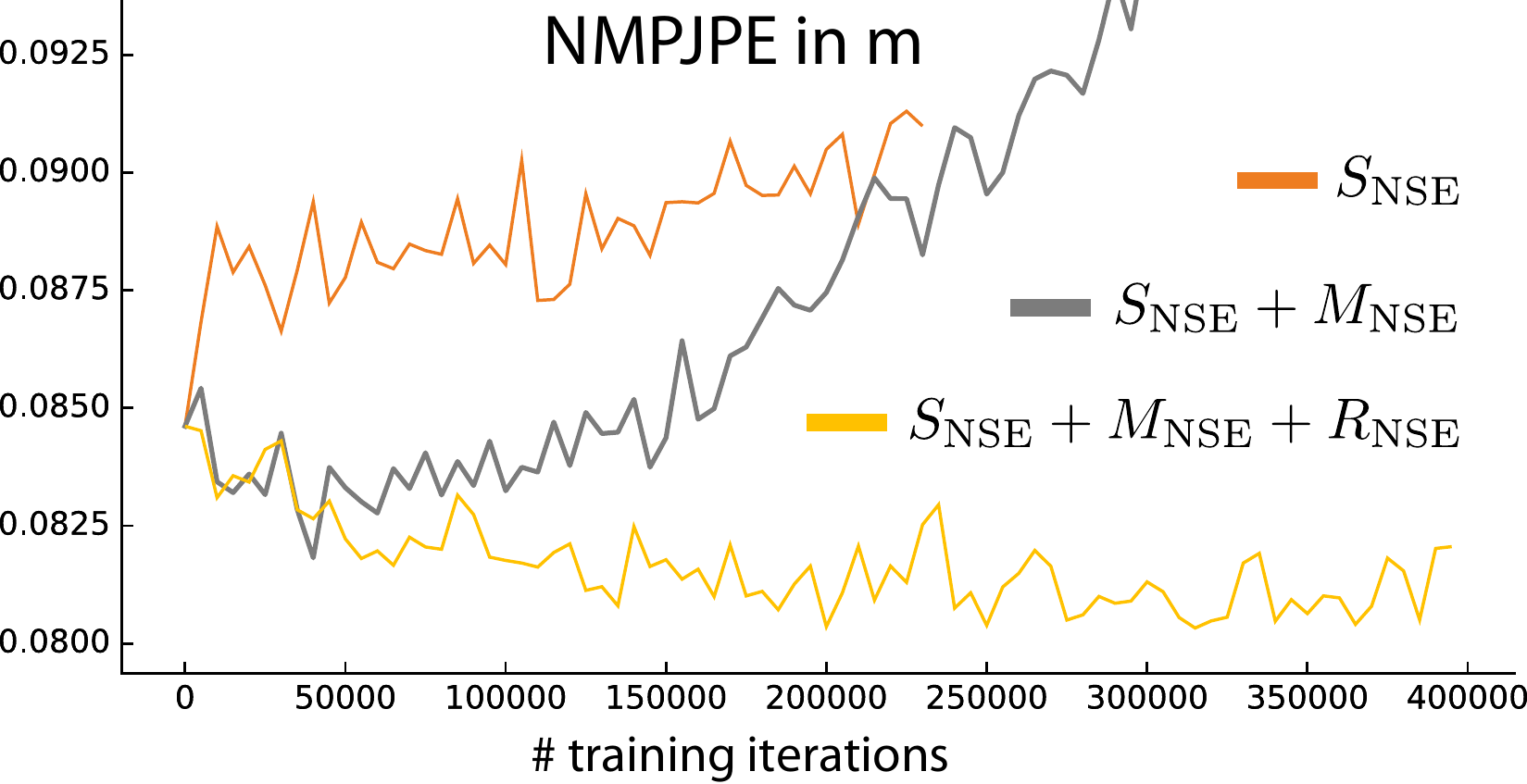}%
	\caption{\label{fig:modelComponentsValidation}%
		{\bf Validation error on H36M.} Mean NMPJPE evaluated at equidistant points during training. 
		With the normalized NSE loss the network performance improves initially, but needs early stopping since it diverges after 30k iterations. In combination with the regularizer $R_\text{NSE}$ the lowest error is attained.
	}
\end{figure}

%% file: fig/h36m1.tex
\begin{figure}
	\centering
	\includegraphics[width=0.9\linewidth]{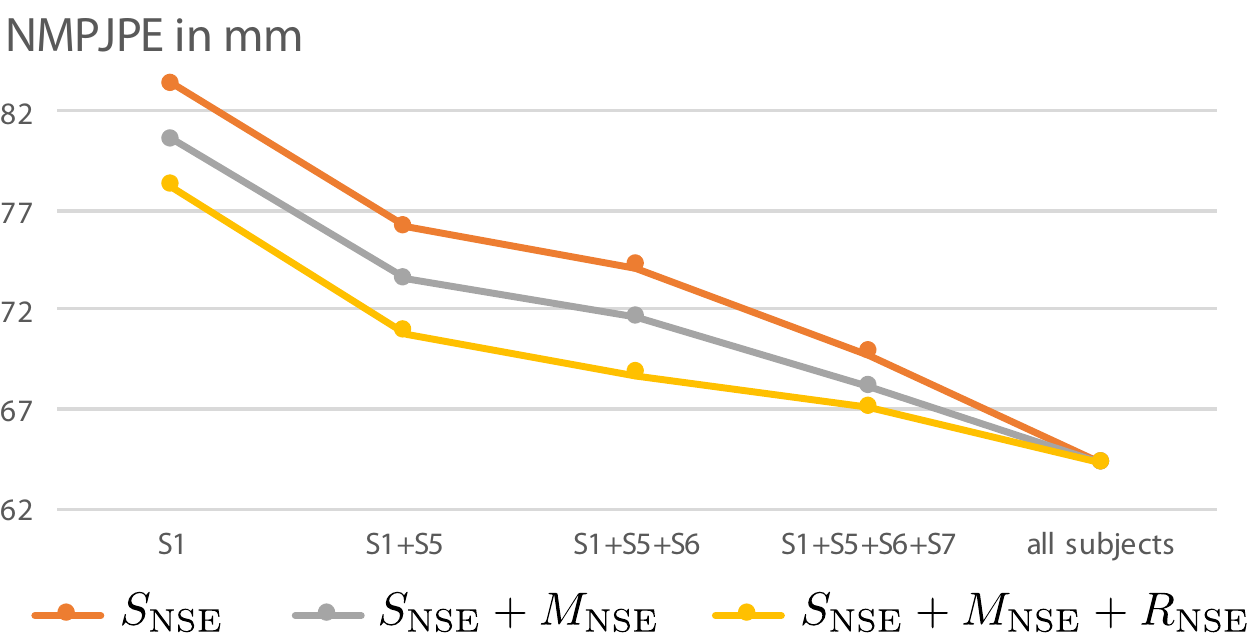}\\
	\caption{\label{fig:h36m_graph}
	Baselines and our full model evaluated on the Human3.6M dataset with varying number of subjects in the labeled set $\cL$. Both, $\OURM$ and $\OURW$ improve consistently.}
\end{figure}

%% file: fig/MPII3DHP.tex
\begin{figure}
	\includegraphics[width=1\linewidth,trim={0 0 0 0},clip]{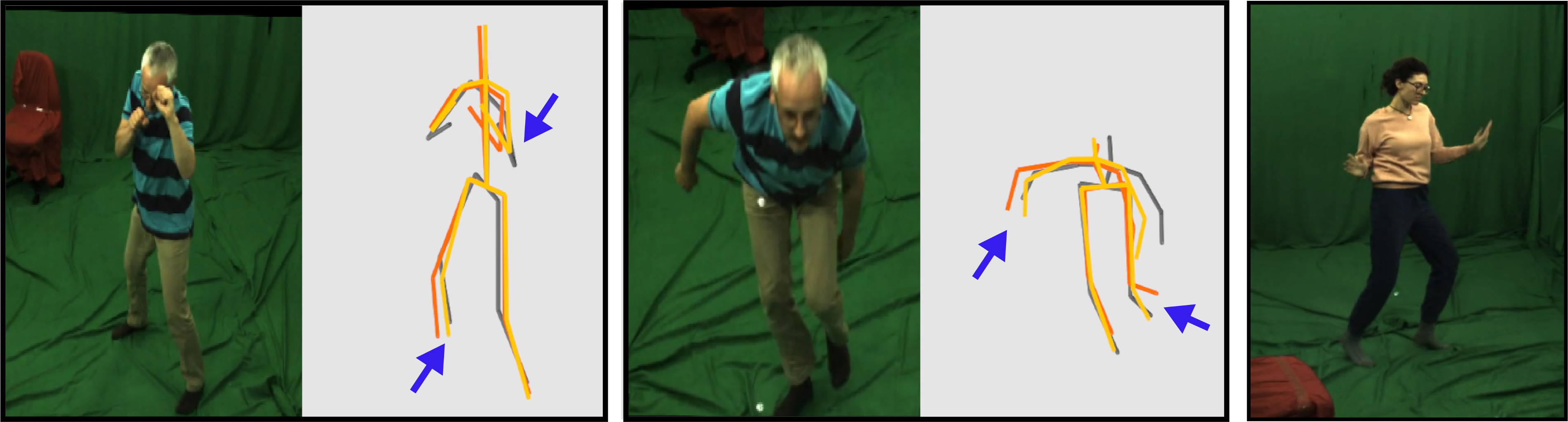}%
	\caption{\label{fig:3DHP_improvements}%
		{\bf Generalization to new camera views.} (left) We overlay \OURW{} predictions in yellow,  \OURN{} predictions in orange and GT in gray in top-down views that are not part of the training set. (right) Only labeled examples acquired by chest-height camera, like the rightmost image, were used for training.}
\end{figure}

%% file: fig/MPII3DHP_table.tex
\begin{table}
	\center
\resizebox{1\columnwidth}{!}{
	\begin{small}
		\begin{tabular}{ |c|l|c|c|c| }
			\multicolumn{5}{c}{Full H36M training, MPII-3DHP test set}\\
			\hline
			Training & Method & NMPJPE in mm & PCK & NPCK \\
			\hline
			\multirow{3}{*}{H36M}
			&Mehta \etal \cite{Mehta17a} & not available & \bf{64.7} & not available \\
			&Zhou \etal \cite{Zhou17a} & not available & not available & \bf{69.2} \\ 
			&\OURN{} & 141.1 & not applicable & 66.9 \\
			\hline
			\multicolumn{5}{c}{}\\
			\multicolumn{5}{c}{Training and test on MPII-3DHP}\\
			\hline
			Training & Method & NMPJPE in mm &PCK & NPCK \\
			\hline
			\multirow{2}{*}{S1,S2,\dots,S8}
			& Mehta \etal \cite{Mehta17a} & not available & $\mathbf{72.5}$ & not available \\
			& \OURN & 101.5 & not applicable & $\mathbf{78.8}$ \\
			\hline
			\multicolumn{5}{c}{}\\
			\multicolumn{5}{c}{Supervised training on MPII-3DHP S1, weakly-supervised on S2 to S8}\\
			\hline
			Training & Method & NMPJPE in mm &PCK & NPCK \\
			\hline
			\multirow{2}{*}{S1, known $\mR$}
			& \OURN & 124.9 & not applicable & 71.6 \\
			& \OURW & \bf{119.8} & not applicable & \bf{73.1} \\
			\hdashline
			{S1, unknown $\mR$} &\OURW{}  & \bf{121.8} & not applicable & \bf{72.7} \\
			\hline
			\multicolumn{5}{c}{}\\
			\multicolumn{5}{c}{Generalization to new viewpoints through weak supervision}\\
			\multicolumn{5}{c}{(labeled S1, views [0, 2, 4, 7, 8]; unlabeled S2 to S7, views 0-9; test on S8, views [1,3,5,6,9])}\\
			\hline
			Training & Method & NMPJPE in mm &PCK & NPCK \\
			\hline
			\multirow{2}{*}{S1, known $\mR$}
			& \OURN& 125.4 & not applicable & 70.9 \\
			& \OURW{}& \bf{108.7} & not applicable & \bf{77.5} \\
			\hline
		\end{tabular}
	\end{small}}
 \vspace{1pt}
	\caption{{\bf Comparison to the state-of-the-art on MPII-3DHP.} (Top 2 tables) In the fully-supervised case, when training either on H36M or on MPII-3DHP, our baselines \OURM{} and \OURN{} yield NPCK (the higher the better) comparable to the state of the art, with a superior performance when exploiting our $\NSE$ metric. (Third table) In the weakly-supervised scenario, with a single labeled subject, \OURW{} provides a boost in accuracy over our baselines. (Fourth table) Our approach can also be used to improve accuracy on new viewpoints without having to rely on any labeled images from these viewpoints.
	}
	\label{tb:3DHP}
\end{table}

%% file: fig/ski_spoerri.tex
\begin{figure}
	\includegraphics[width=1\linewidth,trim={0 0 0 0},clip]{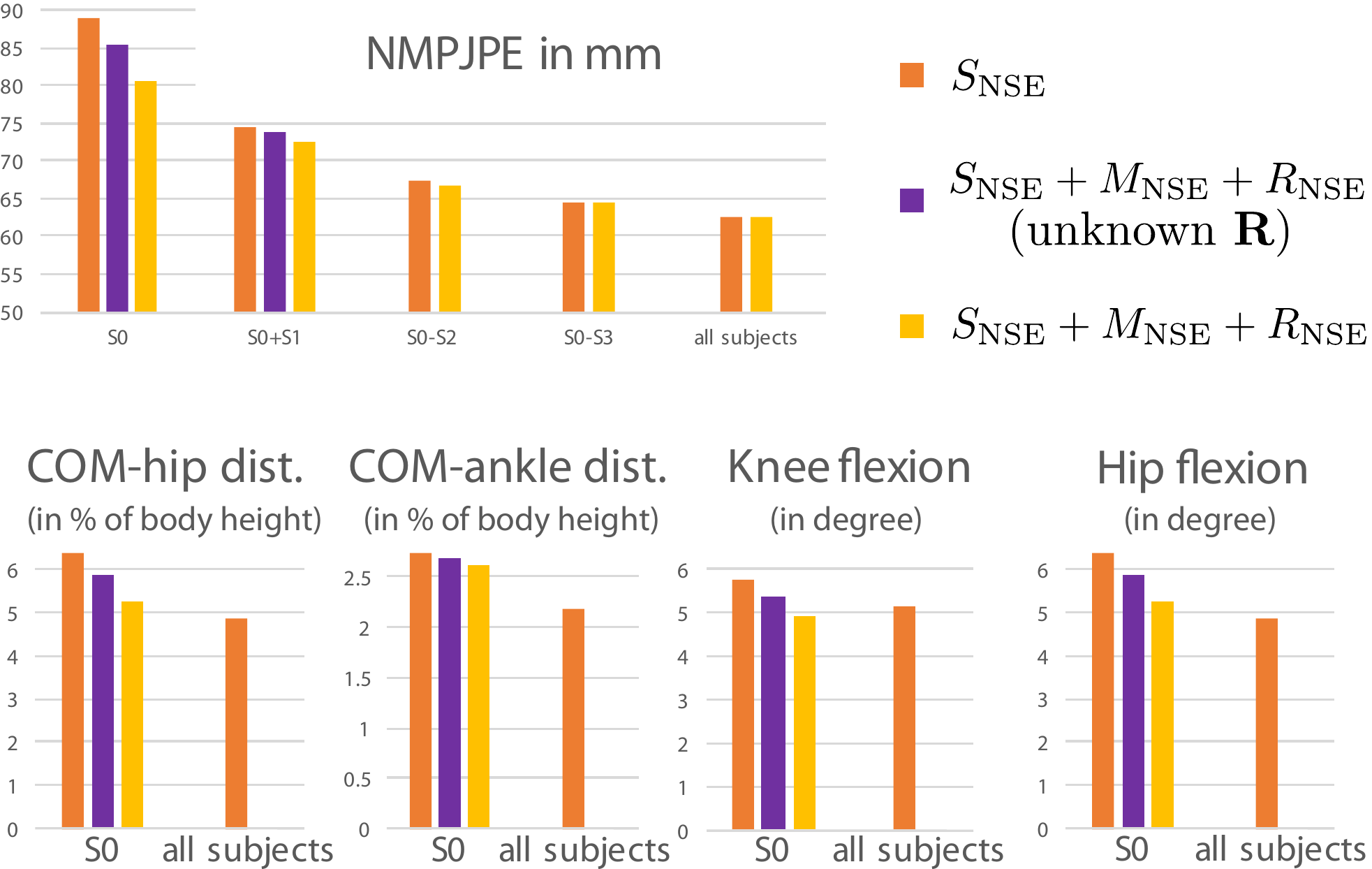}%
	\caption{\label{fig:evalSKISPOERRI}\label{fig:evalSKIMeasurements}%
		{\bf Bar plot of improvements} of \OURW{} in comparison to the baseline \OURN{} on the task of ski-motion reconstruction for varying amount of labeled training data.
		Ski specific measurements, such as the knee flexion angle in degree, are measured alongside the NMPJPE.
	}
\end{figure}

\begin{figure}
	\includegraphics[width=1\linewidth,trim={0 0 0 0},clip]{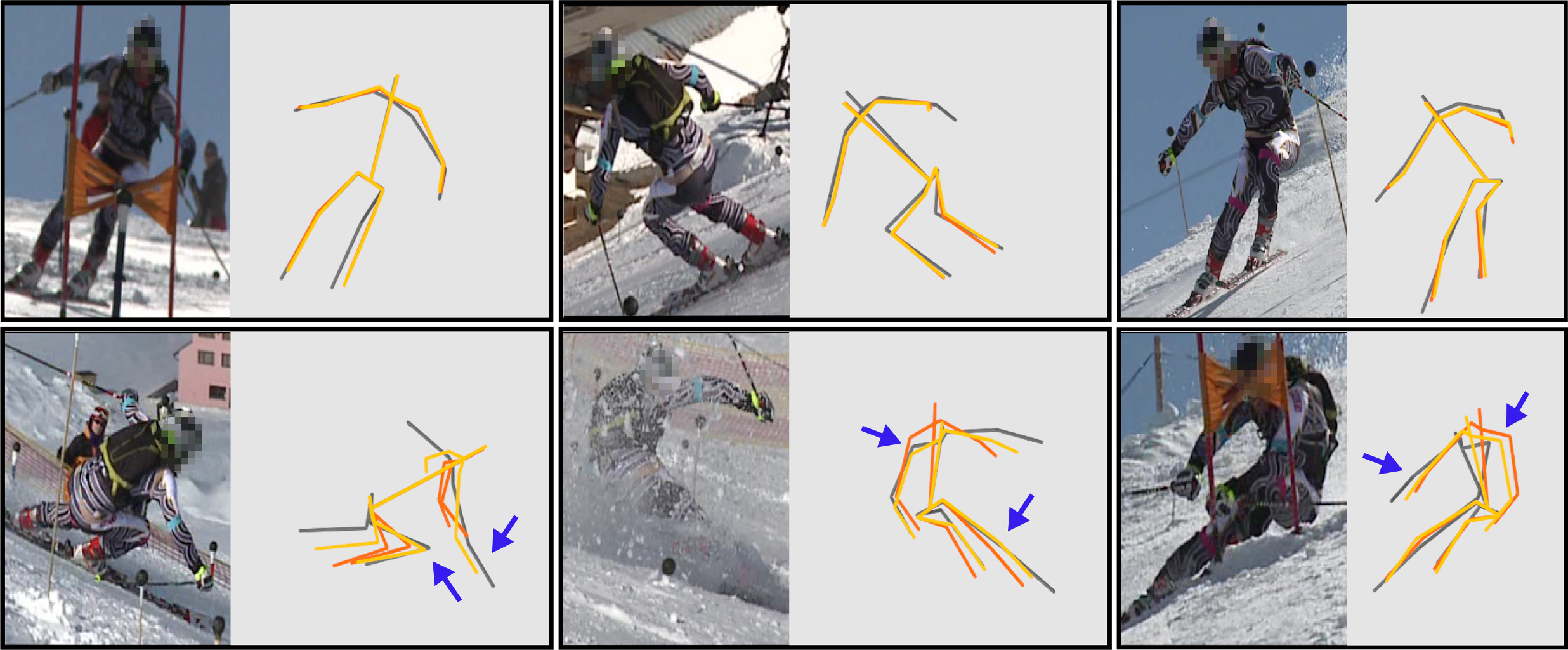}%
	\caption{\label{fig:ski_spoerri_improvements}%
		{\bf Skiing reconstruction.} Training with additional unsupervised data improves reconstruction in difficult cases that are not covered by $\cL$, such as partial occlusion by snow and gate crossings, and extreme poses. We overlay \OURW{} predictions in yellow,  \OURN{} predictions in orange and GT in gray.
	}
\end{figure}

%% file: tex/discussion_conclusion.tex
\section{Conclusion}

We have shown that a small annotated dataset of images and corresponding 3D poses could be effectively supplemented by a set of images acquired by multiple synchronized cameras using minimal amounts of supervision, even if their relative positions are not exactly known.  At the heart of our approach are a multi-view consistency loss that encourages the regressor to predict consistent 3D poses in all views and a regularization loss that prevents the classifier from behaving differently on the annotated images and on the others.
A key limitation of our current approach is that we work on individual images even though we are using video data. A clear next step will be to enforce temporal consistency both on the camera motions and the predictions at training time to increase performance. Our existing multi-view framework should make this relatively straightforward since multiple images acquired over time are not fundamentally different from multiple images acquired at the same time, except for the fact that the pose cannot be expected to be exactly the same.

\comment{
\section{Discussion and Limitations}

We have developed the first approach to learn monocular 3D human pose estimation from unlabeled multi-view images by integrating view-consistency into the training process.
It reduces the number of required annotations and, hence, the manual effort to deal with new motions, and is even applicable when the relative position and orientation between the cameras is unknown.
Specifically, the use of multi-view data was made possible via three core contributions: (i) A robust multi-view consistency term $M$, that inherently performs outlier rejection by relying on a consensus set; (ii) The use of a normalized distance metric \NSE{}, which avoids trivial solutions during optimization;  and (iii) A regularizer $R$ that prevents the multi-view poses to drift to implausible solutions.

Our experiments have demonstrated that our weak supervision consistently improves the results over a model trained on scarce labeled data only. Furthermore, they have shown the importance of our regularizer to both improve prediction accuracy and stabilize the training process. By applying our approach to ski sequences acquired in unconstrained conditions and with rotating cameras, we have evidenced that our algorithm can truly be deployed in the real world, for scenarios where acquiring labeled data is genuinely challenging. Note that our approach is not limited to human pose; it could therefore prove useful to capture animals in their natural habitats, or, potentially, any kind of object motion and deformation.

Nevertheless, the accuracy of our weakly-supervised method remains lower than what would be achieved if all the data were labeled. We believe that unveiling the full potential of weakly-supervised learning requires much larger datasets than the existing ones. Fortunately, the creation of such unlabeled datasets is easier, but nonetheless requires automation to scale by orders of magnitude. The videos need to be synchronized automatically, not only via common visual features~\cite{Elhayek2012}, but, for robustness, by exploiting sound and potentially GPS signal. The subjects must be localized and identified in each view, for which multi-view \cite{Berclaz09a} and monocular methods exist~\cite{Baque17b}, but need to be robustified to work with minimal user intervention.

Altogether, we believe that this paper demonstrates the feasibility and potential of using multiple views as weak supervision for monocular 3D pose estimation. However, much work remains to be done in this area, and we hope that our approach will pave the way for novel methods in this scope, and motivate the collection and processing of massive amounts of unlabeled videos.
}

\comment{
\section{Discussion and Conclusion}

We have developed the first approach to learn monocular 3D human pose estimation from unlabeled multiview images by integrating view-consistency into the training process.
It reduces the number of required annotations and, hence, the manual effort to deal with new motions, and is even applicable when the relative position and orientation between the cameras is unknown.

Nonetheless, the accuracy is higher if the exact camera orientation is known. 
It is a first step towards unsupervised training, but further research is necessary to close the remaining gap to fully supervised methods. %

On the same amount of data, supervised training will always be in favor. To unveil the full potential of unsupervised learning datasets must be much larger than existing sets.
The creation of unlabeled datasets is easier, however, automation is required to scale in magnitudes. %
Videos need to be synchronized automatically, through common visual \cite{Elhayek2012}, sound events, or the GPS signal.
In order to reconstruct camera pose, structure from motion can be applied, but difficulties in crowded scenes with moving backgrounds need to be overcome.
Finally, subjects must be localized and re-identified in each view to enforce consistency, for which multi-view \cite{Berclaz09a} and monocular methods exist \cite{Baque17b}, but need to be advanced to work with minimal user intervention.

In this paper, we focus on algorithmic advances to utilize such data and verify its effectiveness on existing sets where person localization information is available and ground truth can be used for quantitative evaluation. We demonstrate that this route is feasible, which prompts and paves the way for novel methods to collect and process massive amounts of unlabeled multi-view video.

We addressed the problem of capturing motions for which little training examples are available. 
By utilizing consistency constraints between multiple views, at training time {\it only}, we could train the system to predict more accurate 3D human pose from monocular images during test time.
This reduces the amount of manual annotation effort and complements sophisticated deep-net architectures that have to be trained on very large training datasets.
The utilization of multiview examples was made possible through three core contributions.
The robust multi-view consistency term $M$ which relies on outlier rejection through consensus sets. The utilization of the normalized distance metric \NSE{} which avoids trivial solutions during optimization. And third, the regularizer term $R$ that prevents drift towards implausible solutions.
Furthermore, we propose a method to estimate camera pose alongside human pose
in cases where calibration of the camera orientation is intricate, such as for moving pan-tilt-zoom cameras.
We empirically demonstrate that these contributions improve on established benchmarks significantly as well as succeed on on a new Ski dataset with rotating cameras and alpine ski motion.
In sum, succeeding with less annotate examples eases the capture of new motion forms in  environments for which no training data exists. The proposed algorithm is not limited to human pose. It might form an attractive tool for capturing animals in their natural habitats, or, even further, any kind of object motion and deformation.
}

\comment{
	\paragraph{Pose and height} Estimating the height of a person (or object in general) from a single 2D projection is impossible without prior knowledge. Anatomical constraints of the human might reveal some height information, however, it is highly unlikely that this can be learned from the handful of subjects present in the available datasets. Previous work concurs on this argument and uses scale normalization or Procrustes alignment during evaluation, however, we are not aware of studies that utilize the proposed normalization during training by back-propagating the loss gradient through the normalization operation. It can be considered as a generalization of the SoftMax loss on classification tasks. \HR{can one say that?}
}

%% file: top.bbl
\begin{thebibliography}{10}\itemsep=-1pt

\bibitem{Chen16}
W.~Chen, H.~Wang, Y.~Li, H.~Su, Z.~Wang, C.~Tu, D.~Lischinski, D.~Cohen-or, and
  B.~Chen.
\newblock {Synthesizing Training Images for Boosting Human 3D Pose Estimation}.
\newblock In {\em 3DV}, 2016.

\bibitem{Clauser71}
C.~Clauser, J.~McConville, and J.~Young.
\newblock {Weight, Volume, and Center of Mass Segments of the Human Body}.
\newblock {\em Journal of Occupational and Environmental Medicine}, 13(5):270,
  1971.

\bibitem{Fasel18}
B.~Fasel, J.~Sp{\"o}rri, J.~Chardonnens, J.~Kr{\"o}ll, E.~M{\"u}ller, and
  K.~Aminian.
\newblock Joint inertial sensor orientation drift reduction for highly dynamic
  movements.
\newblock {\em IEEE journal of biomedical and health informatics},
  22(1):77--86, 2018.

\bibitem{Fasel16}
B.~Fasel, J.~Sp{\"o}rri, M.~Gilgien, G.~Boffi, J.~Chardonnens, E.~M{\"u}ller,
  and K.~Aminian.
\newblock Three-dimensional body and centre of mass kinematics in alpine ski
  racing using differential gnss and inertial sensors.
\newblock {\em Remote Sensing}, 8(8):671, 2016.

\bibitem{Fischler81}
M.~Fischler and R.~Bolles.
\newblock {Random Sample Consensus: A Paradigm for Model Fitting with
  Applications to Image Analysis and Automated Cartography}.
\newblock {\em Communications ACM}, 24(6):381--395, 1981.

\bibitem{Garg16}
R.~Garg, G.~Carneiro, and I.~Reid.
\newblock {Unsupervised CNN for Single View Depth Estimation: Geometry to the
  Rescue}.
\newblock In {\em European Conference on Computer Vision}, pages 740--756,
  2016.

\bibitem{He16}
K.~He, X.~Zhang, S.~Ren, and J.~Sun.
\newblock {Deep Residual Learning for Image Recognition}.
\newblock In {\em Conference on Computer Vision and Pattern Recognition}, pages
  770--778, 2016.

\bibitem{Horn87}
B.~Horn.
\newblock {Closed-Form Solution of Absolute Orientation Using Unit
  Quaternions}.
\newblock {\em Journal of the Optical Society of America}, 4(4):629--642, April
  1987.

\bibitem{Ionescu14b}
C.~Ionescu, J.~Carreira, and C.~Sminchisescu.
\newblock {Iterated Second-Order Label Sensitive Pooling for 3D Human Pose
  Estimation}.
\newblock In {\em Conference on Computer Vision and Pattern Recognition}, 2014.

\bibitem{Ionescu15}
C.~Ionescu, O.~Vantzos, and C.~Sminchisescu.
\newblock {Matrix backpropagation for Deep Networks with Structured Layers}.
\newblock 2015.

\bibitem{Joo15}
H.~Joo, H.~Liu, L.~Tan, L.~Gui, B.~Nabbe, I.~Matthews, T.~Kanade, S.~Nobuhara,
  and Y.~Sheikh.
\newblock {Panoptic Studio: A Massively Multiview System for Social Motion
  Capture}.
\newblock In {\em International Conference on Computer Vision}, 2015.

\bibitem{Lassner17}
C.~Lassner, J.~Romero, M.~Kiefel, F.~Bogo, M.~Black, and P.~Gehler.
\newblock {Unite the People: Closing the Loop Between 3D and 2D Human
  Representations}.
\newblock In {\em Conference on Computer Vision and Pattern Recognition}, 2017.

\bibitem{Li14a}
S.~Li and A.~Chan.
\newblock {3D Human Pose Estimation from Monocular Images with Deep
  Convolutional Neural Network}.
\newblock In {\em Asian Conference on Computer Vision}, 2014.

\bibitem{Loper15}
M.~Loper, N.~Mahmood, J.~Romero, G.~Pons-Moll, and M.~Black.
\newblock {SMPL: A Skinned Multi-Person Linear Model}.
\newblock {\em ACM SIGGRAPH Asia}, 34(6), 2015.

\bibitem{Martinez17}
J.~Martinez, R.~Hossain, J.~Romero, and J.~Little.
\newblock {A Simple Yet Effective Baseline for 3D Human Pose Estimation}.
\newblock In {\em International Conference on Computer Vision}, 2017.

\bibitem{Mehta17a}
D.~Mehta, H.~Rhodin, D.~Casas, P.~Fua, O.~Sotnychenko, W.~Xu, and C.~Theobalt.
\newblock {Monocular 3D Human Pose Estimation in the Wild Using Improved CNN
  Supervision}.
\newblock In {\em International Conference on 3D Vision}, 2017.

\bibitem{Mehta17b}
D.~Mehta, S.~Sridhar, O.~Sotnychenko, H.~Rhodin, M.~Shafiei, H.~Seidel, W.~Xu,
  D.~Casas, and C.~Theobalt.
\newblock {Vnect: Real-Time 3D Human Pose Estimation with a Single RGB Camera}.
\newblock In {\em ACM SIGGRAPH}, 2017.

\bibitem{Pavlakos16}
G.~Pavlakos, X.~Zhou, K.~Derpanis, G.~Konstantinos, and K.~Daniilidis.
\newblock {Coarse-To-Fine Volumetric Prediction for Single-Image 3D Human
  Pose}.
\newblock In {\em Conference on Computer Vision and Pattern Recognition}, 2017.

\bibitem{Pavlakos17}
G.~Pavlakos, X.~Zhou, K.~D.~G. Konstantinos, and D.~Kostas.
\newblock {Harvesting Multiple Views for Marker-Less 3D Human Pose
  Annotations}.
\newblock In {\em Conference on Computer Vision and Pattern Recognition}, 2017.

\bibitem{Popa17}
A.-I. Popa, M.~Zanfir, and C.~Sminchisescu.
\newblock {Deep Multitask Architecture for Integrated 2D and 3D Human Sensing}.
\newblock In {\em Conference on Computer Vision and Pattern Recognition}, 2017.

\bibitem{Rhodin16}
H.~Rhodin, C.~Richardt, D.~Casas, E.~Insafutdinov, M.~Shafiei, H.-P. Seidel,
  S.~B, and C.~Theobalt.
\newblock {Egocap: Egocentric Marker-Less Motion Capture with Two Fisheye
  Cameras}.
\newblock {\em ACM SIGGRAPH Asia}, 35(6), 2016.

\bibitem{Rogez16}
G.~Rogez and C.~Schmid.
\newblock {Mocap Guided Data Augmentation for 3D Pose Estimation in the Wild}.
\newblock In {\em Advances in Neural Information Processing Systems}, 2016.

\bibitem{Rogez17}
G.~Rogez, P.~Weinzaepfel, and C.~Schmid.
\newblock {Lcr-Net: Localization-Classification-Regression for Human Pose}.
\newblock In {\em Conference on Computer Vision and Pattern Recognition}, 2017.

\bibitem{Simon17}
T.~Simon, H.~Joo, I.~Matthews, and Y.~Sheikh.
\newblock Hand keypoint detection in single images using multiview
  bootstrapping.
\newblock In {\em Conference on Computer Vision and Pattern Recognition},
  volume~2, 2017.

\bibitem{Sun17}
X.~Sun, J.~Shang, S.~Liang, and Y.~Wei.
\newblock Compositional human pose regression.
\newblock In {\em International Conference on Computer Vision}, volume~2, 2017.

\bibitem{Tekin17a}
B.~Tekin, P.~M{\'{a}}rquez{-}neila, M.~Salzmann, and P.~Fua.
\newblock {Learning to Fuse 2D and 3D Image Cues for Monocular Body Pose
  Estimation}.
\newblock In {\em International Conference on Computer Vision}, 2017.

\bibitem{Tome17}
D.~Tome, C.~Russell, and L.~Agapito.
\newblock {Lifting from the Deep: Convolutional 3D Pose Estimation from a
  Single Image}.
\newblock In {\em arXiv preprint, arXiv:1701.00295}, 2017.

\bibitem{Tulsiani17}
S.~Tulsiani, T.~Zhou, A.~Efros, and J.~Malik.
\newblock Multi-view supervision for single-view reconstruction via
  differentiable ray consistency.
\newblock In {\em Conference on Computer Vision and Pattern Recognition},
  volume~1, page~3, 2017.

\bibitem{Varol17}
G.~Varol, J.~Romero, X.~Martin, N.~Mahmood, M.~Black, I.~Laptev, and C.~Schmid.
\newblock Learning from synthetic humans.
\newblock In {\em Conference on Computer Vision and Pattern Recognition}, 2017.

\bibitem{Yan16}
X.~Yan, J.~Yang, E.~Yumer, Y.~Guo, and H.~Lee.
\newblock {Perspective Transformer Nets: Learning Single-View 3D Object
  Reconstruction Without 3D Supervision}.
\newblock In {\em Advances in Neural Information Processing Systems}, pages
  1696--1704. 2016.

\bibitem{Zhou17a}
X.~Zhou, Q.~Huang, X.~Sun, X.~Xue, and Y.~We.
\newblock {Weakly-Supervised Transfer for 3D Human Pose Estimation in the
  Wild}.
\newblock {\em arXiv Preprint}, 2017.

\end{thebibliography}
